\documentclass[letterpaper]{article} % DO NOT CHANGE THIS
\usepackage{aaai24}  % DO NOT CHANGE THIS
\usepackage{times}  % DO NOT CHANGE THIS
\usepackage{helvet}  % DO NOT CHANGE THIS
\usepackage{courier}  % DO NOT CHANGE THIS
\usepackage[hyphens]{url}  % DO NOT CHANGE THIS
\usepackage{graphicx} % DO NOT CHANGE THIS
\usepackage{natbib}  % DO NOT CHANGE THIS AND DO NOT ADD ANY OPTIONS TO IT
\usepackage{caption} % DO NOT CHANGE THIS AND DO NOT ADD ANY OPTIONS TO IT
\usepackage{algorithm}
\usepackage{algorithmic}

\usepackage{newfloat}
\usepackage{listings}
\DeclareCaptionStyle{ruled}{labelfont=normalfont,labelsep=colon,strut=off} % DO NOT CHANGE THIS
\floatstyle{ruled}
\newfloat{listing}{tb}{lst}{}
\floatname{listing}{Listing}
\usepackage{array}
\usepackage{multirow}
\usepackage{tabularx}
\usepackage{amsmath}
\usepackage{xcolor}
\usepackage{amssymb}
\usepackage{mathrsfs}
\newcommand{\Skip}[1]{}

\title{VLCounter: Text-aware Visual Representation for Zero-Shot Object Counting}
\author {
    Seunggu Kang, %\textsuperscript,
    WonJun Moon, %\textsuperscript,
    Euiyeon Kim, %\textsuperscript,
    Jae-Pil Heo %\textsuperscript
    {\footnote{Corresponding author}}
}
\affiliations {
    Sungkyunkwan University\\
    \{seunggu35, wjun0830, keywi9811, jaepilheo\}@g.skku.edu
}

\begin{document}

\maketitle

\begin{abstract}
Zero-Shot Object Counting~(ZSOC) aims to count referred instances of arbitrary classes in a query image without human-annotated exemplars.
To deal with ZSOC, preceding studies proposed a \textbf{two-stage} pipeline: discovering exemplars and counting.
However, there remains a challenge of vulnerability to error propagation of the sequentially designed two-stage process.
In this work, we propose an \textbf{one-stage} baseline, Visual-Language Baseline~(VLBase), exploring the implicit association of the semantic-patch embeddings of CLIP.
Subsequently, we extend the VLBase to Visual-language Counter~(VLCounter) by incorporating three modules devised to tailor VLBase for object counting.
First, we introduce Semantic-conditioned Prompt Tuning~(SPT) within the image encoder to acquire target-highlighted representations.
Second, Learnable Affine Transformation~(LAT) is employed to translate the semantic-patch similarity map to be appropriate for the counting task.
Lastly, we transfer the layer-wisely encoded features to the decoder through Segment-aware Skip Connection~(SaSC) to keep the generalization capability for unseen classes.
Through extensive experiments on FSC147, CARPK, and PUCPR+, we demonstrate the benefits of our end-to-end framework, VLCounter.
Code is available at https://github.com/seunggu0305/VLCounter
\end{abstract}

\section{Introduction}

% OC to ZSOC 등장까지
Object counting, which was initially studied for specific targets, e.g., crowds~\cite{h8}, cells~\cite{2018cell}, animals~\cite{2016animal}, and cars~\cite{2016cars}, has shown that the number of objects can be counted even within a dense image.
Furthermore, recent works have shown significant advances to infer the number of arbitrary objects with several human-annotated exemplar patches.
However, such a strong prerequisite that every cumbersome guidance must be equipped is undoubtedly the main challenge to overcome to grant applicability to object counting methods.
In this context, Zero-Shot Object Counting~(ZSOC) was proposed to mitigate the need for human labor.

\begin{figure}[t]
    \begin{center}
    \includegraphics[width=\linewidth]{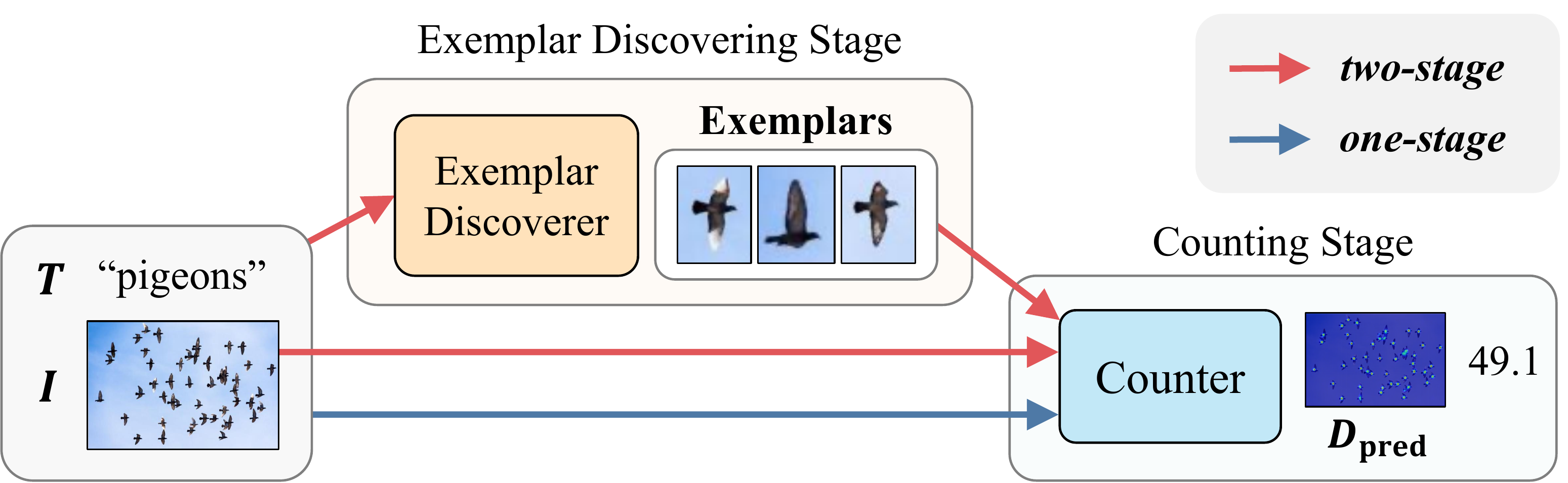}
    \end{center}
    \caption{
        Comparison between two-stage pipeline and one-stage pipeline~(ours).
        The two-stage pipeline requires training the exemplar discoverer~(orange) before the counter~(blue), along with the need for an extra training dataset to optimize the discoverer.
        In contrast, our one-stage pipeline is designed to be simpler and does not necessitate any additional data or training stage. 
    }
    \label{fig:intro}
\end{figure}

% 기존 ZSOC는 2-stage라는 단점을 가짐
Current ZSOC approaches commonly adopt a two-stage pipeline as illustrated in Fig.~\ref{fig:intro}.
These works primarily focus on identifying exemplar patches within the image and subsequently adopt the counting framework from the literature of few-shot object counting~\cite{2022BMNet, 2021FAMNet}.
To identify the exemplar patches, RepRPN~\cite{2022RepRPN} considered the repetition score to detect object patches that frequently appear within the image.
Requirement for counting the desired classes over frequent ones, ZSC~\cite{2023zsc} utilized the class names to enable the class specification.
They localize exemplars by identifying the k-nearest neighbors of the class name embeddings among randomly cropped patches.
Despite their progress, the potential localization error propagation in the two-stage training pipeline~\cite{nag2022zero} is an untapped problem in ZSOC frameworks.
Indeed, they utilized additional datasets to train decent exemplar discovery networks.

% VLBase 설명
This paper pursues a simplified zero-shot object counting framework.
We instantiate an end-to-end ZSOC counter namely Visual-Language Baseline~(VLBase), which consists of a CLIP~\cite{2021clip} encoder and counting decoder.
By leveraging the embedding space of CLIP which enables the implicit association of the semantic and patch embeddings to localize the target object~\cite{2022maskclip, li2023clipsurgery}, VLBase eliminates the need for an exemplar discovery process.

% VLCounter 설명
Additionally, we introduce VLCounter which is built upon VLBase by incorporating three modules devised to tailor VLBase for object counting.
First, we propose Semantic-conditioned Prompt Tuning~(SPT) which extends the visual prompt tuning~(VPT) to efficiently finetune CLIP for the counting task.
Instead of utilizing na\"ive learnable prompts, SPT employs conditioning via semantic embedding to generate patch embeddings that emphasize the region of interest.
Subsequently, based on our observation that the similarity maps between patch embeddings obtained using SPT and semantic embeddings already provide a decent approximation of object locations, we employ simple Learnable Affine Transformation~(LAT) to adjust only the finer details.
Finally, to equip the decoder with the generalization capability and provide rich clues, we exploit intermediate features across different encoding layers of CLIP through Segment-aware Skip Connections~(SaSC).
With all components combined, our simple end-to-end one-stage framework records new state-of-the-art results on the FSC147~\cite{2021FAMNet} dataset validating its superiority over the previous ZSOC methods.
Moreover, we provide additional evidence of cross-dataset generalization by evaluating performance on the car counting dataset CARPK~\cite{2017drone}.

Our contributions are three-fold:
\begin{itemize}
    \item We instantiate an end-to-end baseline for ZSOC, VLBase, by exploiting the vision-language association capability of CLIP. 
    \item We propose a VLCounter consisting of SPT, LAT, and SaSC that allows the model to utilize the generalization capability of CLIP in a counting-specific manner.
    \item Our experiments on FSC147 and cross-dataset validation verify the effectiveness of VLCounter.
\end{itemize}
\section{Related Works}
% 08/06 03:00 - WJ : Class-specific, Zero-shot fix
%% TODO : Few-shot, Learnable Prompt Tuning

\subsection{Object Counting}
\paragraph{Class-specific Object Counting}
focuses on quantifying specific class samples, e.g., crowds~\cite{h0, h1, h2, h3}, cars~\cite{2016cars, 2017drone}, animals~\cite{2016animal}, and cells~\cite{2018cell}.
Most works fall into two main categories each employing detection~\cite{d0, 2017drone, d1} or regression~\cite{r0, r1, r2, r3} mechanism to measure the number of instances.
The former predicts the bounding box for every instance using an object detector, whereas the latter predicts the density distribution of the image instead, thereby being recognized as a more robust stream against partially occluded objects~\cite{2019GMN}.

\paragraph{Few-shot Object Counting}
To overcome the lack of generality of being constrained to a specific class, 
Generic Matching Network~(GMN)~\cite{2019GMN} first formalized class-agnostic object counting to count the desired objects provided by the human-annotated exemplar patches.
They introduced a two-stream architecture to encode each image and exemplar to handle the difference in their resolution. 
Following them, CFOCNet~\cite{2021CFOCNet} and BMNet~\cite{2022BMNet} also adopted and enhanced the two-stream approach by adding a layer-wise matching procedure and bilinear similarity metric.
Other works adhere to single-stream architecture. 
To be specific, FamNet~\cite{2021FAMNet} and RCAC~\cite{gong2022class} use ROI pooling after feature extraction to obtain exemplar prototypes.
However, the aforementioned studies suffer from the limitation that every inference requires human-annotated exemplars.

\paragraph{Zero-shot Object Counting}
has been proposed by RepRPN~\cite{2022RepRPN} to discard the duty of annotating target exemplars for counting.
To be specific, they trained the region proposal network~(RPN) to capture the patches containing the most frequently appeared objects to replace human-annotated exemplars.
Then, to further grant more applicability to exemplar-free object counter, ZSC~\cite{2023zsc} presented a method that takes guidance from semantic information.
By matching semantic information to randomly generated patches, they sampled the most semantically relevant patches to obtain target exemplars.
Our work shares the goal with ZSC in that we aim to train the counter that can count user-specified classes with only class names.
Yet, as mentioned methods adopt a two-stage pipeline that is prone to error propagation, we focus on mitigating such issues by proposing an end-to-end framework that localizes and counts at once.

\subsection{Prompt Tuning}
\label{Related:VPT}
Prompt tuning is a popular strategy to adapt pre-trained large models for downstream tasks due to its efficiency compared to conventional fine-tuning methods~\cite{wang2022learning, gu2021ppt, 2020lp, 2022vpt}.
Whereas fine-tuning updates all parameters, prompt tuning freezes the pre-trained large models and introduces only a small set of learnable prompts to optimize~\cite{li2021prefix, 2022vpt}.
Following these works, we utilize prompt tuning to efficiently exploit the quality of the visual-language understanding capability of pre-trained CLIP.
Yet, our work differs in using semantic information from the semantic embeddings to condition the prompts in the visual encoder to concentrate more on specification-relevant information.
% 전체 overview 그림
\begin{figure*}[t]
    \begin{center}
    \includegraphics[width=\linewidth]{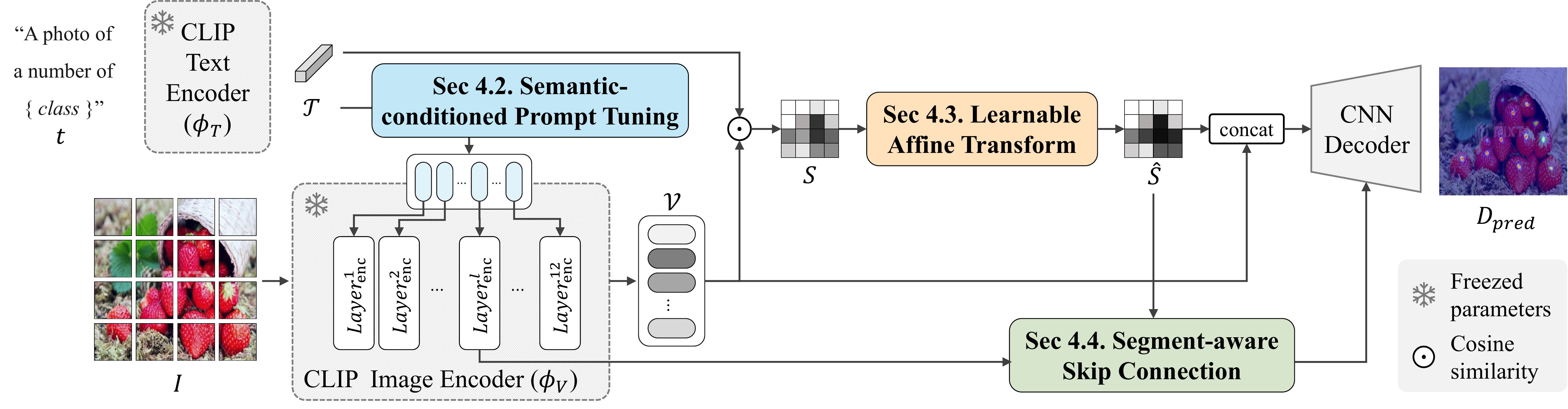}
    \end{center}
    \caption{
        Overview of VLBase and VLCounter: each without and with colored components.
        The end-to-end baseline, VLBase, employs CLIP encoders to extract both image and text embeddings. 
        Then, the decoder processes the image-text similarity map along with visual embeddings to count the number of specified objects.
        With three colored modules, VLCounter leverages the generalization capability of VLBase to be tailored for object counting.
    }
    \label{fig:VLCounter_overview}
\end{figure*}

\section{Preliminaries}
\subsection{Problem Formulation: ZSOC}
\label{Method:Formulation}
ZSOC aims to predict the density map $D\in \mathbb{R}^{H\times W \times 1}$ for image $I\in \mathbb{R}^{H\times W \times3}$ that belongs to unseen classes $C^u$~($f:(I, C^u)\mapsto D$) without any visual exemplar clues.
In the training stage, the model is trained with $\mathcal{D}_\text{train} = \{(I_i, C^s_i, D_i)\}_{i=1}^{i=\mathbb{N}}$ where $C^s_i$ denotes the seen class names during training.
Then in the testing stage, the model is to yield a density map for $\mathcal{D}_\text{test}=\{(I_i, C^u_i, D_i)\}_{i=\mathbb{N}+1}^{i=\mathbb{M}}$, where $C^s \cap C^u=\varnothing$.

\subsection{Overview of CLIP}
\label{Method:preliminaries}
This section introduces the underlying motivation behind our proposed method.
CLIP is composed of two encoders: an image encoder $\phi_{V}(\cdot)$ and a text encoder $\phi_{T}(\cdot)$.
The text encoder takes prompted class name~$t$ e.g., \textit{A photo of [$kiwi$]} and produces a semantic embedding $\mathcal{T} \in \mathbb{R}^{1 \times d}$ where $d$ represents an embedding dimension.
The image encoder takes a learnable class token $[cls]$ along with embedded patch sequences $V$ as inputs and encodes global and local semantics in the class token $[cls]$ and patch tokens $\mathcal{V}$ respectively.
Note that $V = [v_1, v_2,...,v_N] \in \mathbb{R}^{N \times (P^2 \cdot d)}$ where $N$ is the number of embedded patches, and $(P \cdot P)$ is the resolution of each patch.
Formally, this process can be expressed as follows:
\begin{equation}
    \mathcal{T} = \phi_{T}(t); \quad\quad[\text{ }[cls],\mathcal{V}\text{ }] = \phi_{V}([\text{ }[cls], V\text{ }]).
\end{equation}
These encoders are trained collaboratively to map $\mathcal{T}$ and $[cls]$ into a shared representation space.

Recently, there exist studies suggesting the implicit localization capability of CLIP, where each patch embedding preserves local image semantics~\cite{2022maskclip, li2023clipsurgery}.
And this property, coupled with the powerful image-text joint embedding space of CLIP, has provided a clear motivation for utilizing CLIP as a robust tool for zero-shot segmentation~(localization).~\cite{li2022languagedriven, rao2022denseclip, luddecke2022image}.
Taking similar inspiration yet focused on object counting, we aim to leverage the implicit localization capability of CLIP to achieve precise and efficient object counting in an end-to-end manner.

\section{Visual-Language Counter: End-to-End Framework for Zero-Shot Object Counting}
This section presents Visual-Language Counter~(VLCounter), an efficient end-to-end ZSOC framework.
We first establish a baseline model referred to as Vision-Language Baseline (VLBase), which exploits the visual-language localization capacity of CLIP in Sec.~\ref{Method:VLBase}.
Then, we bring three improvements on top of VLBase to introduce VLCounter.
Specifically, we emphasize the regions of interests~(Sec.~\ref{Method:SPT}), learn task-specific visual-language similarity~(Sec.~\ref{Method:LAT}), and exploit semantic-relevant information across the multi-level representations~(Sec.~\ref{Method:SaSC}).
The overall architectures of the two models are illustrated in Fig.~\ref{fig:VLCounter_overview}.

\subsection{Visual-Language Baseline}
\label{Method:VLBase}
VLBase is a standalone baseline, eliminating the need for few-shot counting techniques that previous ZSOC approaches heavily rely on.
Given input query image $I$ and class name $C$, VLBase obtains patch embedding $\mathcal{V}$ and semantic embedding $\mathcal{T}$ using CLIP encoders $\phi_V(\cdot)$ and $\phi_T(\cdot)$, respectively.
By calculating the cosine similarity between $\mathcal{T}$ and $\mathcal{V}$, the similarity map $S\in \mathbb{R}^{H \times W}$ is yielded:
\begin{equation}
    S_{ij}(\mathcal{V},\mathcal{T}) = \frac{v_{ij}\mathcal{T}^\mathsf{T}}{||v_{ij}||||\mathcal{T}||},
\end{equation}
where $S_{ij}$ corresponds to the value at position $(i,j)$ in matrix $S$ and $v_{ij}$ represents the embedding at position $(i, j)$ of 2D-reshaped $\mathcal{V}$.

As mentioned in prior studies~\cite{2022maskclip, li2023clipsurgery}, we observed that the similarity map between CLIP-encoded semantic and patch embeddings provides an adequate indication of the degree of semantic similarity between the patch and semantic embedding.
We find that this similarity map is a decent clue for a decoder to localize the target objects.
Consequently, the CNN-based counting decoder predicts the density map $D_\text{pred}$ by utilizing features of $\mathcal{V}$ and $S$:
\begin{equation}
\label{eq3}
    D_\text{pred} = \phi_\text{decoder}([\mathcal{V}, S]),
\end{equation}
where $[\cdot,\cdot]$ denotes channel-wise concatenation.
Finally, the object count prediction is derived by summing all values in $D_\text{pred}$.

\paragraph{Counting Loss}
For training, we adopt a conventional MSE loss:
\begin{equation}
\label{eq3countingloss}
    \mathcal{L}_\text{count} = ||D_\text{pred} - D_\text{gt} ||^2_2,
\end{equation}
where $D_\text{gt}$ denotes the ground truth density map.

% SPT 모듈 그림
\begin{figure}[h]
    \begin{center}
    \includegraphics[width=\linewidth]{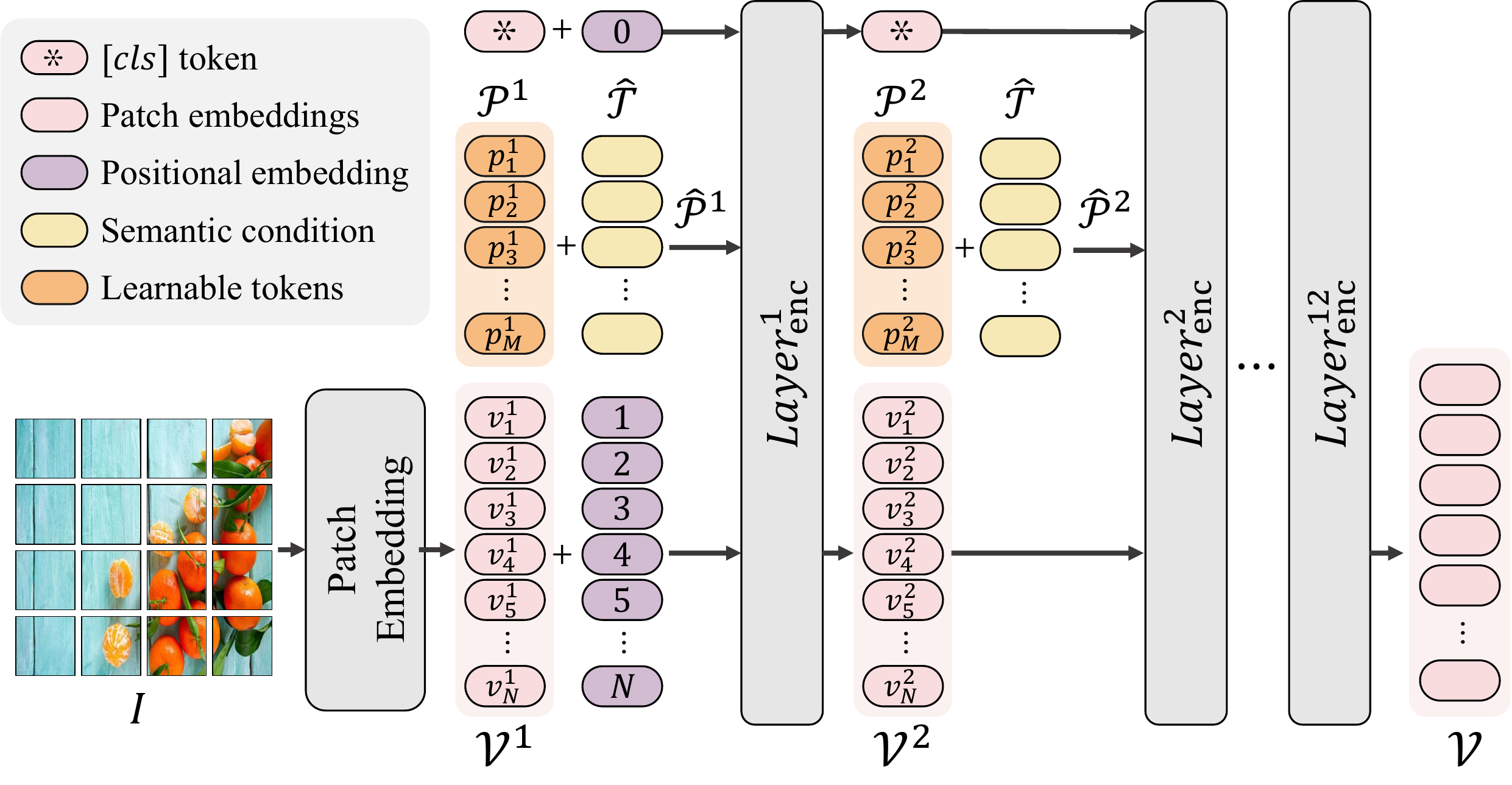}
    \end{center}
    \caption{
        Illustration for Semantic-conditioned Prompt Tuning~(SPT).
        In addition to learnable visual prompts~(orange) in the image encoder, text features~(yellow) are integrated to specify the desired semantics.
    }
    \label{fig:SPT}
\end{figure}

\subsection{Semantic-conditioned Prompt Tuning~(SPT)}
\label{Method:SPT}
To grant task-specificity to the CLIP image encoder without sacrificing its generalization capability, a straightforward approach is to employ visual prompt tuning~(VPT)~\cite{2022vpt}.
However, the na\"ive VPT, which simply concatenates a few learnable tokens to the input sequence of each encoding layer does not take the semantic information into account.
Hence, we introduce Semantic-conditioned Prompt Tuning~(SPT), which utilizes semantic information along with the learnable tokens to assist the image encoder to extract target-semantic-highlighted visual features.

Specifically, as illustrated in Fig.~\ref{fig:SPT}, SPT has new learnable tokens for each encoding layer.
Learnable tokens for $l^{th}$ layer are defined as $\mathcal{P}^l = [p^l_1, p^l_2, ..., p^l_M]$ where the number of learnable tokens is denoted as $M$.
These tokens are then, supplemented with the linearly projected semantic embedding $\hat{\mathcal{T}}$ to generate semantic-conditioned prompts $\hat{\mathcal{P}}$.
The semantic-conditioned prompts for the $l^{th}$ layer are defined as follows:
\begin{equation}
    \hat{\mathcal{P}}^l = [p_1^l + \hat{\mathcal{T}}, p_2^l + \hat{\mathcal{T}}, p_M^l + \hat{\mathcal{T}}],
\end{equation}
where $\hat{\mathcal{T}} = \phi_c(\mathcal{T})$ and $\phi_c$ denotes the parameters of the projection layer.
Consequently, with the conditioned prompts $\hat{\mathcal{P}}$, the patch embedding process in $l^{th}$ layer of the image encoder can be expressed as:
\begin{equation}
    [\text{ }[cls],~\rule{0.25cm}{0.15mm},~\mathcal{V}^{l+1}\text{ }] = Layer^{l}_{\text{enc}}([\text{ }[cls],~\hat{\mathcal{P}}^{l},~\mathcal{V}^{l}\text{ }]),
\end{equation}
where initial input $\mathcal{V}^1=[v^1_1, v^1_2, \cdots, v^1_N]$ is a sequence of embedded patches through the patch embedding layer prior to the encoder.
Be aware that we follow VPT~\cite{2022vpt} to discard output tokens of $\hat{\mathcal{P}}$~(represented as $\rule{0.25cm}{0.15mm}$) and do not propagate to the subsequent layer.

\subsection{Learnable Affine Transformation~(LAT)}
\label{Method:LAT}
Through the adoption of the SPT, we obtain visual representations in which the corresponding regions of the target class are highlighted.
Nevertheless, due to the nature of object counting, discovering the central points of the objects rather than encompassing the entire object area, a discrepancy might arise between the information contained in the similarity map $S$ and the loss that needs to be backpropagated during training.

In light of this, we propose learnable affine transformation matrix~(LAT) to facilitate the conversion of similarity map $S$ to counting map $\hat{S}$ and establish a more task-specific visual-semantic linkage centered around individual objects as follows: 
\begin{equation}
    \hat{S} = W \otimes S + B,
\end{equation}
where $W, B \in \mathbb{R}^{H \times W}$ are learnable matrices for affine transformation and $\otimes$ indicates element-wise multiplication.
In addition, we directly optimize the counting map $\hat{S}$ with the rank-aware contrastive loss to learn the proper degree of activation for object counting. 
Details of rank-aware contrastive loss are elaborated in Sec.~\ref{Method:Loss}.
With LAT, the input to the decoder $[\mathcal{V}, S]$ in Eq.~\ref{eq3} of VLBase is replaced by $[\mathcal{V}, \hat{S}]$.

% SaSC 모듈 그림
\begin{figure}[t!]
    \begin{center}
    \includegraphics[width=\linewidth]{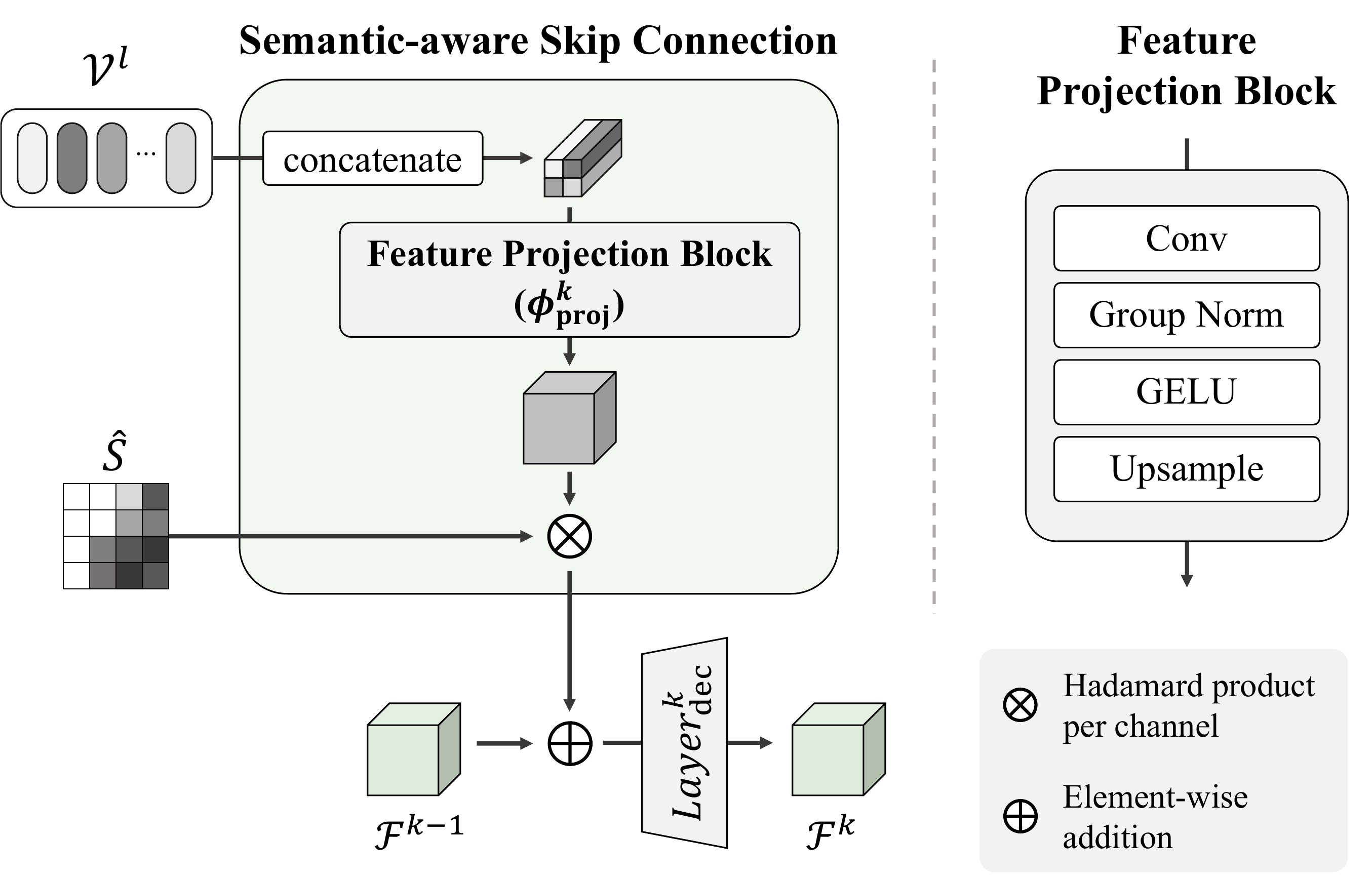}
    \end{center}
    \caption{
        The flow of Semantic-aware Skip Connection~(SaSC) and architecture of feature projection block.
        Intermediate visual features are projected and filtered with an object-aware counting map $\hat{S}$ to produce object-relevant encoder features. Consequently, these are integrated into its counterpart in the decoder.
    }
    \label{fig:SaSC}
\end{figure}

\subsection{Segment-aware Skip Connection~(SaSC)}
\label{Method:SaSC}
For ZSOC, where the model encounters unseen classes during inference, it is important to train a decoder that is tailored for object counting while maintaining a generalization ability.
Sharing the motivation with VLBase in Sec.~\ref{Method:VLBase} that CLIP features inherently preserve local semantics, we adopt skip connections that incorporate intermediate features of the encoder to its counterpart in the decoder.

\begin{table*}[t]
    \small
    \centering
    \setlength\tabcolsep{0pt}
    \setlength{\extrarowheight}{2.3pt}
    
    \begin{tabular*}{\textwidth}{@{\extracolsep{\fill}}*{1}{l} @{\extracolsep{\fill}}*{8}{c}}
    \hline

    % Header
    \multirow{2}{*}{Methods} & \multirow{2}{*}{Stage} & Class & \multirow{2}{*}{Train Dataset} & \multicolumn{2}{c}{Val set} & \multicolumn{2}{c}{Test set} & Inference \\%\multirow{2}{*}{inf.speed} \\
    \cline{5-6}\cline{7-8}
    & & Name & & MAE~$\downarrow$ & RMSE~$\downarrow$ & MAE~$\downarrow$ & RMSE~$\downarrow$ & speed~(s)~$\downarrow$\\
    \hline

    % Few-shot
    \textbf{\textit{few-shot}} & & & & & & \\
    GMN~\cite{2019GMN} & 1 & $\times$ & FSC147 & 29.66 & 89.81 & 26.52 & 124.57 & - \\
    FamNet~\cite{2021FAMNet} & 1 & $\times$ &FSC147 & 24.32 & 70.94 & 22.56 & 101.54 & 0.82 \\
    BMNet~\cite{2022BMNet} & 1 & $\times$ & FSC147 & 19.06 & 67.95 & 16.71 & 103.31 & 0.86 \\
    BMNet+~\cite{2022BMNet} & 1 & $\times$ & FSC147 & 15.74 & 58.53 & 14.62 & 91.83 & 1.59 \\
    \hline

    % Zero-shot
    \textbf{\textit{zero-shot}} & & & & & & & \\
    RepRPN-Counter~\cite{2022RepRPN} & 2 & $\times$ & FSC147~+~MS COCO~
    & 31.69 & 100.31 & 28.32 & 128.76 & - \\
    ZSC~\cite{2023zsc} & 2 & \checkmark & FSC147~+~MS COCO & 26.93 & 88.63 & 22.09 & 115.17 & 0.86+$\alpha$ \\
    VLBase~(Ours) & 1 & \checkmark & FSC147 & 31.82 & 98.89 & 32.20 & 130.51 & 0.81 \\
    VLCounter~(Ours) & 1 & \checkmark & FSC147 & \textbf{18.06} & \textbf{65.13} & \textbf{17.05} & \textbf{106.16} & 0.82 \\
    \hline
    
    \end{tabular*}
    \caption{
        Quantitative comparison to state-of-the-art approaches on the FSC147 dataset. 
        % GT stands for human-annotated boxes and Patch-Selection refers to selected patches with ZSC~\cite{2023zsc} approach.
        $\alpha$ in the rightmost column indicates an additional cost necessary for the exemplar discovery process in the context of the two-stage pipeline.
    }
    \label{tab:main}
\end{table*}

As shown in Fig.~\ref{fig:SaSC}, the $l^{th}$ encoder patch features are spatially concatenated and projected to yield decoder-assistive representations.
Then, we multiply the affine transformed similarity map $C$ to emphasize the object-relevant patches.
Finally, these patch features are added to the corresponding $k^{th}$ layer features of the decoder.
Formally, the $k^{th}$ decoding layer with SaSC, receiving $l^{th}$ encoder features, operates as follows:
\begin{equation}
    \mathcal{F}^{k} = Layer^{k}_\text{dec}(\mathcal{F}^{k-1} + \phi_\text{proj}^{k}(\mathcal{V}^{l}) \otimes \hat{S}),
\end{equation}
where $\phi_\text{proj}^{k}(\cdot)$, $\mathcal{F}^{k}$, and $\otimes$ stand for the parameter of feature projection block, the output of the $k$-th decoding layer, and Hadamard products per channel, respectively.

\subsection{Training Objectives}
\label{Method:Loss}
\Skip{
    To facilitate precise local visual-language alignment between patch embedding and semantic embedding
}
In addition to the counting loss described in Eq.~\ref{eq3countingloss}, VLCounter additionally employs rank-aware contrastive loss~\cite{hoffmann2022ranking, moon2023query}.
Whereas the $\mathcal{L}_{\text{count}}$ trains the whole model to learn the counting objective, 
our focus in SPT and LAT is learning to yield the counting-tailored similarity map in the encoder.
In this regard, we adopt rank-aware contrastive loss in the counting map $\hat{S}$ to assign higher activations on the patches that are nearby the object centers.
To design the hierarchical guidance for a rank-aware contrastive loss, we first normalize the ground truth density map $D_{\text{gt}}$ to be mapped between 0 and 1.
Then, we iterate the batch for $K$ times with different thresholds to prepare positive and negative sets; patches are gathered as positive if the corresponding patch in $D_\text{gt}$ has a higher value than the threshold, and if not, as negative.
Formally, the rank contrastive loss with the positive set $\hat{S}_r^\text{pos}$ and the negative set $\hat{S}_r^\text{neg}$ is formulated as follows:
\begin{eqnarray}
    \label{eq7contrastiveloss}
    \mathcal{L}_{\text{rank}}=-\sum_{k=1}^{K}\text{log}\frac{
    \sum_{\hat{S}_i\in \hat{S}_r^\text{pos}}\text{exp}(\hat{S}_i/\tau)}
    {
    \sum_{\hat{S}_j\in (\hat{S}_r^\text{pos} \cup \hat{S}_r^\text{neg})}\text{exp}(\hat{S}_j/\tau)
    },
\end{eqnarray}
where $\tau$ is a temperature scaling parameter.

With the objectives in Eq.~\ref{eq3countingloss} and Eq.~\ref{eq7contrastiveloss} combined, VLCounter's final objective is as follows:
\begin{equation}
    \mathcal{L}_{\text{total}} = \mathcal{L}_{\text{count}} + \lambda \cdot \mathcal{L}_{\text{rank}},
\end{equation}
where $\lambda$ is a hyperparameter to balance between the losses.
% CARPK 성능 테이블
\begin{table}[t!]
    \small
    \setlength{\extrarowheight}{2.3pt}
    \setlength{\tabcolsep}{1.5pt}
    \centering
    
    \begin{tabular*}{\linewidth}{l@{\extracolsep{\fill}}*{4}{c}}
    
    \hline
    \multirow{2}{*}{Methods} & \multicolumn{2}{c}{CARPK} & \multicolumn{2}{c}{PUCPR+} \\
    \cline{2-3}\cline{4-5}
    & MAE & RMSE & MAE & RMSE \\
    
    % Few-shot
    \hline
    \textbf{\textit{few-shot}} & & \\
    FamNet & 28.84 & 44.47 & 87.54 & 117.68 \\
    BMNet & 14.61 & 24.60 & 103.18 & 112.42 \\
    BMNet+ & 10.44 & 13.77 & 62.42 & 81.74  \\
    
    % Zero-shot
    \hline
    \textbf{\textit{zero-shot}} \\
    VLBase & 20.47 & 24.33 & 90.82 & 104.01 \\
    VLCounter & \textbf{6.46} & \textbf{8.68} & \textbf{48.94} & \textbf{69.08} \\
    \hline
    
    \end{tabular*}
    
    \caption{Cross-dataset validation performance on the CARPK and PUCPR+ dataset.}
    \label{tab:CARPK}
\end{table}
\section{Experiments}

In this section, we provide a comprehensive explanation of experimental details.
First, we delve into the implementation details, datasets, and evaluation metrics in Sec.~5.1, followed by a comparison of our model with existing state-of-the-art methods in Sec.~5.2.
Then, we conduct an in-depth exploration of each component further in Sec.~5.3.

\subsection{Experimental Details}

\paragraph{Implementation Details.}
For all experiments, we employed CLIP ViT-B/16 as our encoders followed by a decoder consisting of 4 repeated units.
Each of these units consists of one feature projection block in Fig.~\ref{fig:SaSC} and one additional convolutional layer.
Regarding the image input, each image is resized to $384\times384$, and augmentations such as gaussian noise, gaussian blur, horizontal flip, and color jittering were applied.
We trained the model using AdamW~\cite{2017adamw} optimizer with a learning rate of $1\mathrm{e}^{-4}$ and weight decay of $1\mathrm{e}^{-2}$ for 200 epochs with a batch size of 16 on a single NVIDIA RTX A6000.
For temperature scaling and loss-balancing hyperparameter $\lambda$ and $\tau$, we used $1\mathrm{e}^{-6}$ and 1.

\paragraph{Datasets.}
To explore the counting capability of models, we use FSC147~\cite{2021FAMNet}, the first large-scale dataset for class-agnostic counting.
It includes 6135 images from 147 categories mainly composed of foods, animals, kitchen utensils, and vehicles.
We also utilize CARPK and PUCPR+~\cite{2017drone} datasets.
These datasets exhibit different properties from the images in FSC147, so we use them for cross-dataset validation which is to test the model's generality.
To be specific, CARPK consists of 1,448 parking lot images with nearly 90,000 cars taken in a drone view at 40 meters height on average.
On the other hand, PUCPR+ contains nearly 16,456 cars in total which have 10th-floor-view images.

\begin{table}[t]
    \small
    \setlength{\extrarowheight}{2.3pt}
    \setlength{\tabcolsep}{1.5pt}
    \centering
    
    \begin{tabular*}{3.3in}{@{\extracolsep{\fill}}*{8}{c}}

    % Valid
    \hline
    \multirow{2}{*}{No.} & \multirow{2}{*}{SPT} & \multirow{2}{*}{LAT} & \multirow{2}{*}{SaSC} & \multicolumn{2}{c}{Val set} & \multicolumn{2}{c}{Test set} \\
    \cline{5-6}\cline{7-8}
    & & & & MAE & RMSE & MAE & RMSE \\
    \hline
    M1 & $\times$ & $\times$ & $\times$ & 31.82 & 98.89 & 32.20 & 130.51 \\
    M2 & \checkmark & $\times$ & $\times$ & 20.61 & 75.36 & 17.58 & 112.89 \\
    M3 & $\times$ & \checkmark & $\times$ & 29.97 & 96.59 & 28.26 & 127.44 \\
    M4 & $\times$ & $\times$ & \checkmark & 24.88 & 81.28 & 24.16 & 113.01 \\
    M5 & \checkmark & \checkmark & \checkmark & \textbf{18.06} & \textbf{65.13} & \textbf{17.05} & \textbf{106.16} \\
    \hline
    
    \end{tabular*}    
    \caption{  
        Ablation study on each component of VLCounter.
        % Ablation study to analyze the components of VLCounter. 
        % Ablation study on learnable affine transform~(LAT), text conditioned prompt tuning~(TPT), and similarity-aware skip connection~(SaSC).
    }
    \label{tab:ablation}
\end{table}

% \paragraph{Evaluation Metrics.} 
% For a fair comparison, we adopt Mean Average Error~(MAE) and Root Mean Squared Error~(RMSE) to evaluate the performances following the conventions of previous works~\cite{2022BMNet, 2021FAMNet, 2023zsc, 2022RepRPN}.

\subsection{Comparison with State-of-the-art Methods}
We compare VLBase and VLCounter against previous class-agnostic counting methods in Tab.~\ref{tab:main}.
Despite its simple design, the performances of VLBase are comparable to the two-stage methods that even utilize additional training data.
% For VLBase, although its simplicity in its design choice and pipeline, we find its performances are comparable to the two-stage methods that utilize additional training data.
On the other hand, VLcounter clearly surpasses other ZSOC baselines.
Particularly, when compared to ZSC, VLCounter achieves a relative improvement of 32.94\% and 22.81\% in terms of validation MAE and test MAE, respectively.
Moreover, we remark on the comparable results to the state-of-the-art few-shot counting method: BMNet.
This is an especially notable milestone for ZSOC since few-shot methods are generally seen as the upper bound of two-stage ZSOC methods; the counting framework in two-stage works is usually adopted from few-shot methods.

On the rightmost columns, we provide the inference speed per image.
% On the rightmost columns, we mention the number of learnable parameters and inference speed per image.
As our one-stage approaches~(VLBase and VLCounter) only require the time to count the objects, it is shown that their inference speeds are much f   aster than a two-stage method~(ZSC) which needs extra time to discover exemplars~(denoted as $\alpha$ since the implementation is not fully publicized).
In addition to the inference time, VLBase and VLCounter have much fewer parameters to learn, having their strength in shorter training time~(Training time for VLCounter is approximately 2$\times$ faster than BMNet+).

Following previous class-agnostic counting methods~\cite{2021FAMNet, 2022BMNet}, we verify the generalization capability of VLBase and VLCounter by conducting a cross-dataset evaluation on CARPK and PUCPR+ datasets in Tab.~\ref{tab:CARPK}, and VLBase and VLCounter demonstrate their benefits in generalization.
Whereas the performance gaps between few-shot methods and VLBase is reduced, we observe the superiority of VLCounter to other methods by boosting MAE up to 38.12\% and 27.54\% in CARPK and PUCPR+ datasets compared to BMNet+.
In particular, we emphasize the single-digit results of VLCounter in terms of both MAE and RMSE are derived without any fine-tuning~(The average number of cars in each image of CARPK is 62).
We attribute such success in cross-dataset validation to adapting the generality of CLIP to counting-specific and incorporating multi-level features to provide rich semantics into the prediction, each approximately taking 54\% and 46\% portions in the increase in CARPK MAE.

\subsection{Ablation Studies on VLCounter}
\paragraph{Component Analysis.}
To validate the effectiveness of individual components, we conducted an ablation study as presented in Tab.~\ref{tab:ablation}.
% To validate the effectiveness of individual components, we conducted an ablation study in Tab.~\ref{tab:ablation}.
Starting with VLBase~(M1), we add SPT, LAT, and SaSC in M2, M3, and M4, respectively.
Among the individual components, the effectiveness of SPT demonstrated in M2 is the most pronounced.
% The effectiveness of SPT shown in M2 is the best of the individual component. 
This significant improvement demonstrates the importance of fine-tuning incorporated with the semantic condition.
LAT in M3 is another important component.
While it can be seen as not incurring a dramatic increase in performance, the counting map $\hat{S}$ derived from LAT is also an essential element in SaSC.
Lastly, M4 shows that SaSC not only boosts generalization capability but also task-specific predictions.
This is because layer-wise intermediate representations in CLIP encoder are also semantically meaningful~\cite{li2023clipsurgery} and SaSC aggregates them to aid counting prediction.

% 컴포넌트 Ablation
\begin{table}[t]
    \small
    \setlength{\extrarowheight}{2.3pt}
    \setlength{\tabcolsep}{1.5pt}
    \centering
    \begin{tabular*}{\linewidth}{l@{\extracolsep{\fill}}*{5}{c}}
    \hline
    \multicolumn{2}{c}{\multirow{2}{*}{Condition}} & \multicolumn{2}{c}{Val set} & \multicolumn{2}{c}{Test set} \\
    \cline{3-4}\cline{5-6}
     & & MAE & RMSE & MAE & RMSE \\
    \hline
    \multicolumn{2}{c}{VLCounter} & 18.06 & 65.13 & 17.05 & 106.16 \\ \hline
    \textbf{SPT} & w/o $\mathcal{T}'$ & 19.07 & 65.72 & 17.19 & 107.54 \\
    % w/ $\mathcal{T}'$ & 18.06 & 65.13 & 17.05 & 106.16 \\
    \hline
    \textbf{SaSC} & w/o $\hat{S}$ & 20.28 & 65.54 & 19.38 & 105.69 \\
    % \textbf{SaSC} & w/ $S$ & 18.30 & 61.60 & 17.72 & 105.19 \\
    % w/ $C$ & 18.06 & 65.13 & 17.05 & 106.16\\
    \hline
    % \hline
    \end{tabular*}
    \caption{Analysis of semantic-conditioning techniques in SPT and SaSC.}
    \label{tab:ablation_component}
\end{table} 

\begin{table}[t]
    \small
    \setlength{\extrarowheight}{2.3pt}
    \setlength{\tabcolsep}{6pt}
    \centering
    % \begin{tabular}{ccccc}
    \begin{tabular*}{0.9\linewidth}{@{\extracolsep{\fill}}*{5}{c}}

    \hline
    \multirow{2}{*}{Text prompts} & \multicolumn{2}{c}{Val set} & \multicolumn{2}{c}{Test set} \\
    \cline{2-3}\cline{4-5}
     & MAE & RMSE & MAE & RMSE \\
    \hline
    Singular & 20.08 & 67.92 & 19.18 & 105.04 \\
    \hline
    Plural & 18.06 & 65.13 & 17.05 & 106.16 \\ 
    \hline
    \end{tabular*}
    \caption{Analysis of pluralized context to prompt the class names.}
    \label{tab:ablation_plural}
\end{table} 

\paragraph{Effect of conditioning semantic information.}
%In this subsection, 
We further conduct ablation studies on semantic conditioning.
In Tab.~\ref{tab:ablation_component}, we compare conventional VPT with SPT and test the semantic conditioning in SaSC.
Along with the benefits of VPT of granting task-specificity, utilizing semantic conditions in VPT allows the prompts to be more semantically specific.
In addition, using semantic conditions in filtering the knowledge that is passed to the decoder with residual paths clearly benefits SaSC. 
We think that the semantic conditioning with the counting map $\hat{S}$ suppresses the object-irrelevant information, thereby contributing to the improvements.

% 정성 평가 결과 그림
\begin{figure*}[t]
    \begin{center}
        \includegraphics[width=0.95\linewidth]{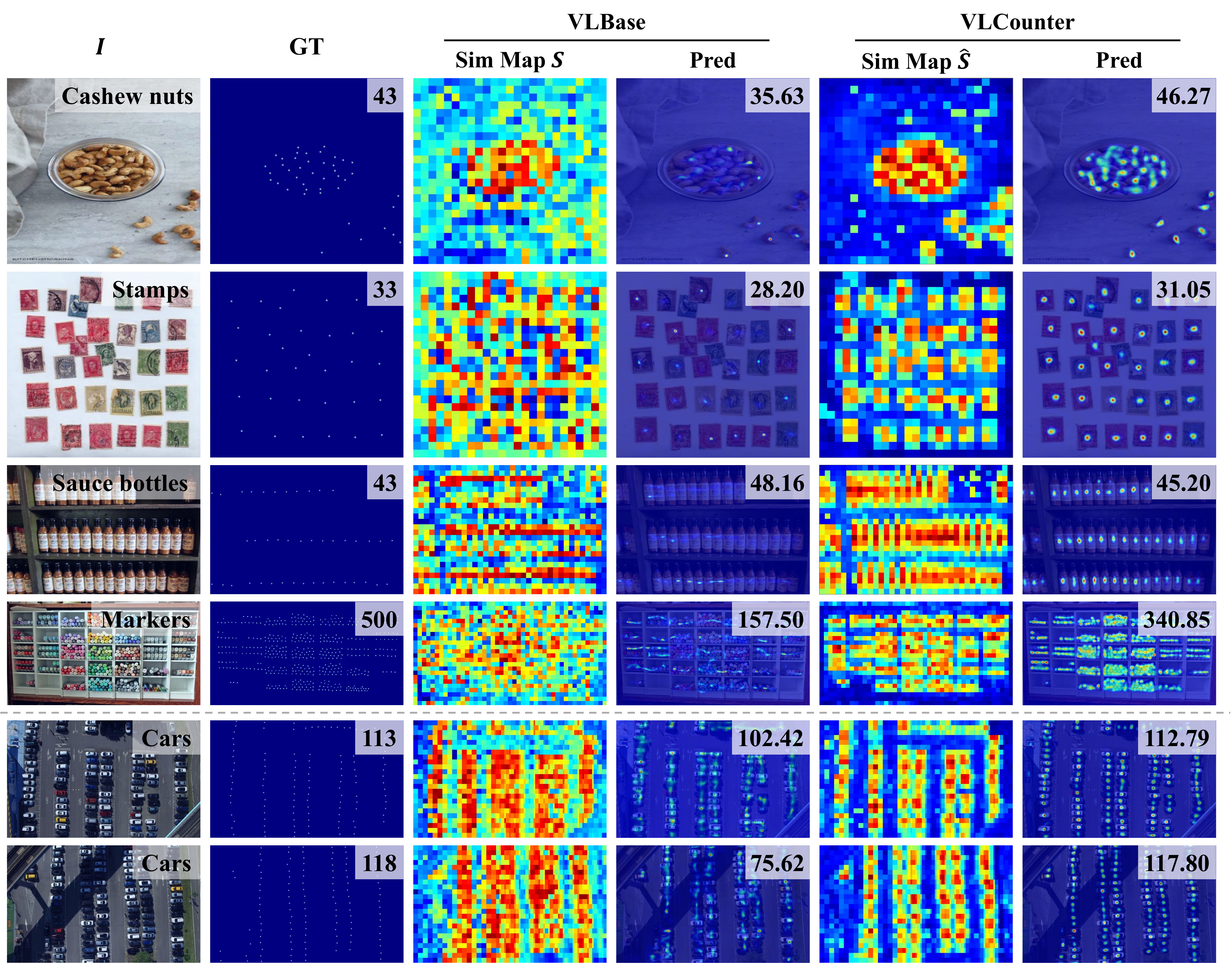}
    \end{center}
    \caption{
        Qualitative comparison of VLBase and VLCounter on the FSC-147~(Top 4 rows) and CARPK~(Bottom 2 rows).
        Class names and counting values are shown at the right top of the query image~($I$) and the predicted density map, respectively.
    }
    \label{fig:qualitative_results}
\end{figure*}

\paragraph{Effect of plural text prompts.}
We followed CLIP~\cite{2021clip} to use different context prompts to encode the semantic embeddings.
Yet, since the counting task mainly assumes the existence of multiple instances in every image, we modified text prompts to be in plural form.
In Tab.~\ref{tab:ablation_plural}, we compare the results between using singular and plural forms of text prompts, and text prompts in plural form have the advantage in the counting task.

\subsection{Qualitative Results}
Along with the quantitative results, we study how the components of VLCounter affect class-specificity.
In Fig.~\ref{fig:qualitative_results}, we compare both the similarity map and the density map of VLBase and VLCounter.
By delivering the semantic condition and fine-tuning the similarity map, we find the similarity map to retain more compact salient regions; the activations in the background are suppressed~(1st, 2nd rows) and object regions are clearly localized~(2nd, 3rd rows).
Then, by aggregating multi-level representations of rich semantics with these similarity maps in the decoder, we observe the clear discrepancy between the predicted density maps from VLBase and VLCounter, especially for densely populated images~(4th row).

Furthermore, we provide the cross-dataset results in the last two rows in Fig.~\ref{fig:qualitative_results}.
Similar to what we discussed with predictions for FSC147, we verify that VLCounter is a counting-tailored and generalizable model across new categories, shapes, and densities of objects.
These results verify the advantage of employing a pretrained vision-language model for capturing the semantics of newly seen objects, i.e., cars.
% With.... and SaSC that passes generalization capability to the decoder via multiple connections, we validate the robustness of VLCounter across new categories, shapes, sizes, and densities of objects.
Refer to the appendix for more visualizations.

\section{Conclusion}
In this work, we present a simple end-to-end framework VLBase and VLCounter for zero-shot object counting that eliminates the need for the process of discovering exemplars.
Simply put, VLBase is built upon the pre-trained vision-language CLIP model.
Then, VLCounter introduces three key components that bring task-specificity and object-specificity.
Whereas the semantic-conditioned prompt tuning and learnable affine transformation fine-tune the encoding process to obtain counting-tailored representations, the segment-aware skip connection is designed to learn the generalizable decoder with the knowledge.
Our thorough experiments on FSC147 and cross-dataset benchmarks validate the effectiveness and efficiency of VLCounter.
% and intuitive \textbf{one-stage} zero-shot object counting method VLBase and VLCounter that builds upon the pre-trained vision-language CLIP models.
% Especially VLCounter involves three key components that further embody the localization capability of the visual-text interaction to suit the object-counting task while obtaining semantic-aware visual representations in an end-to-end manner
% To validate the efficacy of VLCounter, we conduct extensive evaluations on the FSC147 and CARPK benchmark datasets and show that our model outperforms previous state-of-the-art complicated two-stage methods.

% As we focused on designing and fine-tuning the effective end-to-end framework, 
% Although we have successfully achieved 
% From a technical standpoint, we successfully achieved direct interaction between visual and language modalities.
% We did not incorporate considerations for scale variations, as exemplar patches that could serve as references were not available to us.
% As a result, the model might encounter challenges in handling objects at different scales effectively.
% Addressing this issue and incorporating scale considerations could be an area of improvement for future work.

\bibliography{aaai24}

\clearpage
\appendix

\section{Additional Implementation Details}
As indicated in the manuscript, we adopted CLIP ViT-B/16 as our encoders.
Yet, since the image encoder was trained with images of 224x224 resolution, we resized the position embeddings of the image encoder to adapt CLIP to handle the images of 384x384 resolution.
For $\phi_c(\cdot)$ that projects semantic vectors for SPT, we share one linear layer across all visual encoder layers of CLIP.
Also, $M$, the number of learnable tokens in SPT is designated to 10.
For LAT, the learnable matrices $W$ and $B$ are initialized to 1 and 0, respectively, and the thresholds for the rank-contrastive loss are set to [0.8, 0.6, 0.4], establishing the iteration count $K$ to 3.
Finally, for SaSC, we extract encoder feature $\mathcal{V}$ from layers $l = [7,8,9]$, and pass onto the decoder feature $\mathcal{F}$ at layers $k = [2,3,4]$.

\section{Effect of Learnable Tokens}

\subsection{Effect of the Number}
\begin{figure}[ht]
    \begin{center}
    \includegraphics[width=\linewidth]{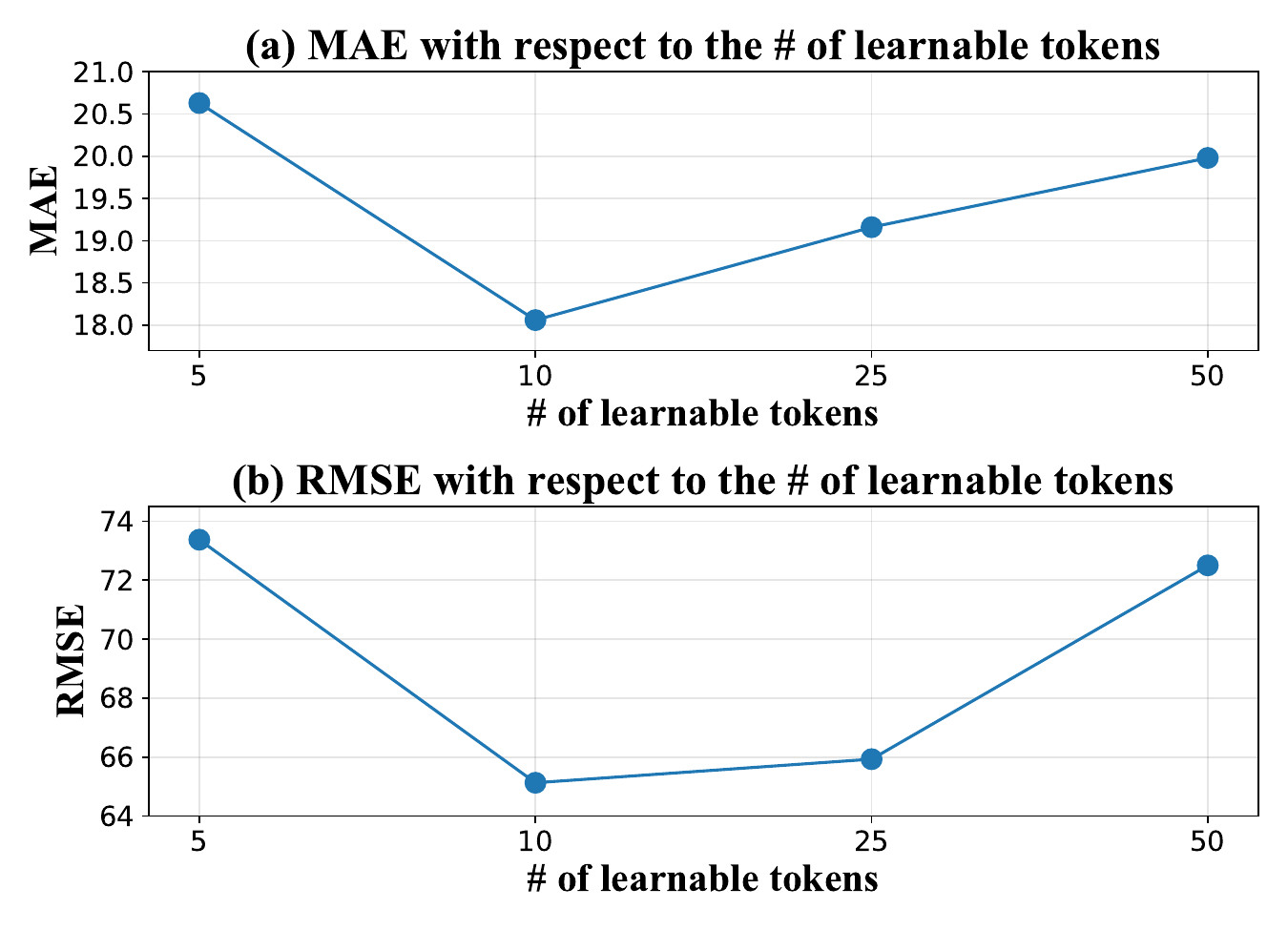}
    \end{center}
    \caption{
        Effect of the number of learnable tokens.
    }
    \label{fig:num}
\end{figure}

We determine the optimal number of learnable tokens required to facilitate an effective transfer of CLIP.
An extremely small number of learnable tokens might not be sufficient to effectively facilitate the transfer of a pre-trained large model.
However, employing an excessive number of visual prompts also can have a detrimental impact on our model's performance due to the loss of generality of CLIP.
Based on experiments as reported in Fig.~\ref{fig:num}, we have determined that the optimal number of learnable tokens for our specific task is 10.

\subsection{Effect of the Depth}
In addition to the apparent influence of the number of learnable tokens on ZSOC performance, we also anticipate that the specific placement of these tokens within the encoder layers will have a substantial impact.
To provide clearer context, we assign numerical labels to the 12 layers of the vision transformer in the CLIP image encoder, ranging from 1 to 12.
We observe that introducing prompt tokens on the earlier layers typically results in improved performance compared to placement on the latter layers as reported in Tab.~\ref{tab:depth}.
The highest performance is achieved when learnable prompt tokens are inserted into every image encoding layer~(layer 1--12), which also serves as the default setting in our experimental setup.

\begin{table}[h]
    \small
    \setlength{\extrarowheight}{2.3pt}
    \setlength{\tabcolsep}{1.5pt}
    \centering
    \begin{tabular*}{\linewidth}{l@{\extracolsep{\fill}}*{5}{c}}
    \hline
    \multicolumn{2}{c}{\multirow{2}{*}{Condition}} & \multicolumn{2}{c}{Val set} & \multicolumn{2}{c}{Test set} \\
    \cline{3-4}\cline{5-6}
     & & MAE & RMSE & MAE & RMSE \\
    \hline
    \multicolumn{2}{c}{1 -- 3} & 20.5 & 68.81 &	19.73 & 111.6 \\
    \multicolumn{2}{c}{1 -- 6} & 19.72 & 66.70 & 20.23 & \textbf{103.44} \\
    \multicolumn{2}{c}{10 -- 12} & 23.34 &	81.39 &	21.81 &	110.92 \\
    \multicolumn{2}{c}{7 -- 12} & 23.32 &	80.05 &	22.16 &	105.42 \\
    \multicolumn{2}{c}{1 -- 12} & \textbf{18.06} & \textbf{65.13} & \textbf{17.05} & 106.16 \\ 
    \hline
    % \hline
    \end{tabular*}
    \caption{Effect of the depth of learnable tokens}
    \label{tab:depth}
\end{table}

\section{LAT Value Distribution}
\begin{figure}[ht]
    \begin{center}
    \includegraphics[width=\linewidth]{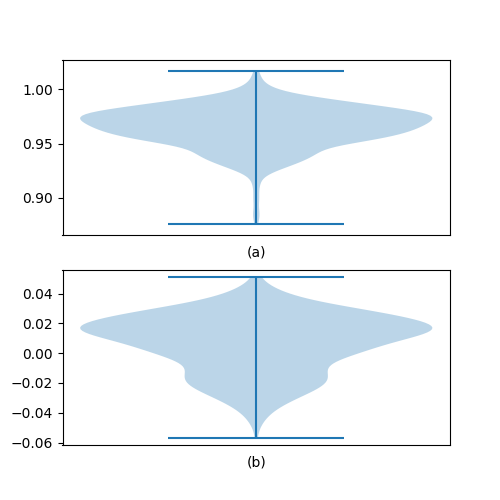}

    \end{center}
    \caption{
        (a) displays a distribution diagram of the values in $W$, while (b) illustrates the distribution of values in $B$, both of which are employed for the affine transformation of LAT.
    }
    \label{fig:LAT}
\end{figure}

In our manuscript, we mentioned that LAT is to facilitate the conversion of similarity maps to be more counting-specific: guiding activations to be more compact on object centers and not significantly modifying the similarity map.
To substantiate our argument, we plot the distributions of $W$ and $B$ matrices in Fig.~\ref{fig:LAT}.
As we observe that the values of $W$ and $B$ are concentrated around 1 and 0, respectively, we confirm that LAT maintains the localization capability of our encoder and only fine-tunes the similarity map to be more counting-specific.

\section{Effect of Encoder Features in SaSC}
Building upon the arguments presented in SaSC, which emphasize the attainment of generalizability and rich semantics through the aggregation of encoder features during decoding, we explore the combinations of the successive layers to yield the best results.
Through this investigation, we aim to determine which layer's features are most conducive to enhancing the overall performance of the decoding process.
\begin{figure}[ht]
    \begin{center}
    \includegraphics[width=\linewidth]{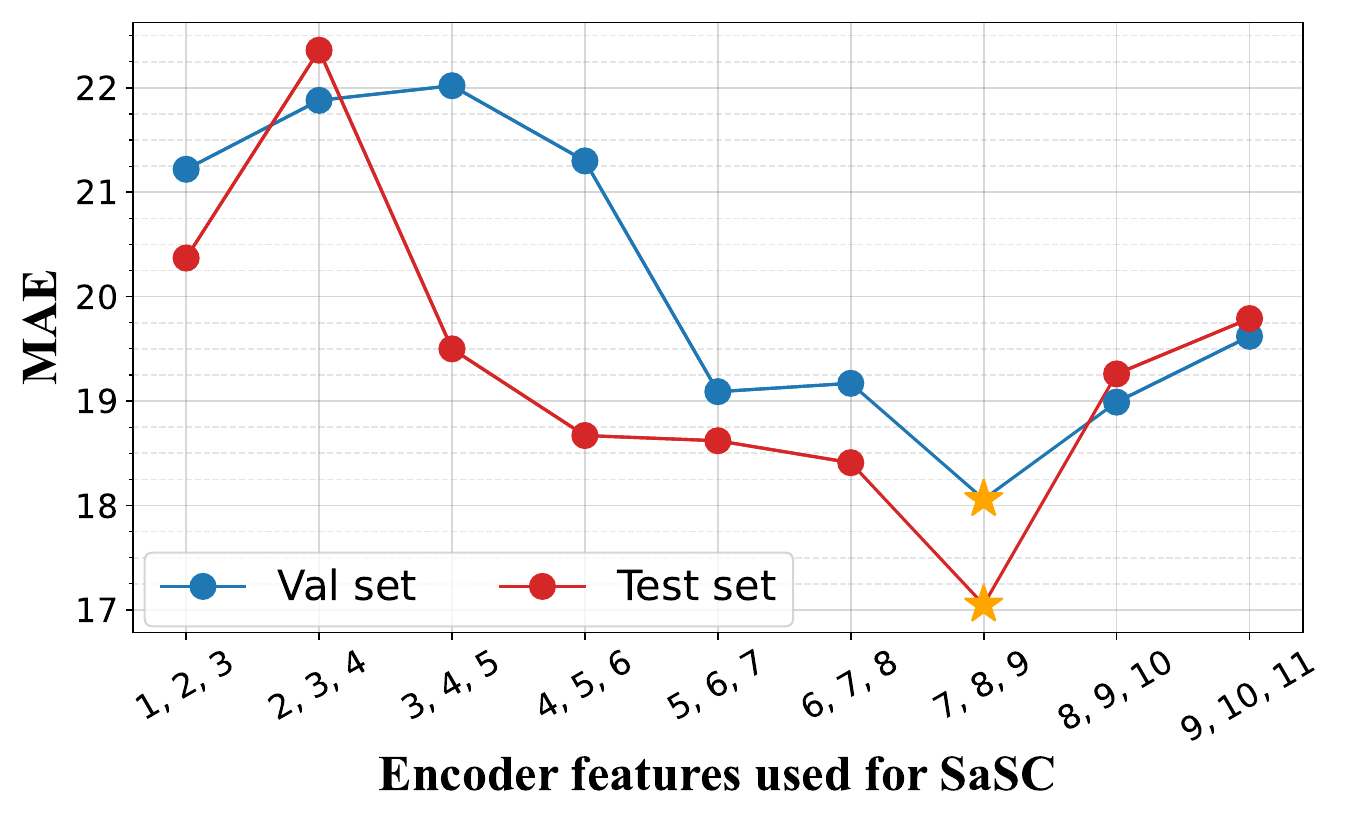}
    \end{center}
    \caption{
        Effect of the combinations of encoder layers.
    }
    \label{fig:sasc}
\end{figure}

Evidently, Fig.~\ref{fig:sasc} shows that the shallow encoder layers do not perform well due to their limited acquisition of meaningful patch-level information.
In addition, we find a similar tendency to the arguments presented by \cite{li2023clipsurgery}, mentioning that  Feed Forward Networks~(FFNs) in the deeper CLIP layers are more likely to bring negative impacts on the vision-language alignment and localization capabilities.
In this context, we have chosen the features from the 7th, 8th, and 9th encoding layers and incorporated them into the 2nd, 3rd, and 4th decoding layers.

\section{Context Prompts}
In Tab.~5 in our manuscript, we demonstrated the influence of the form of context prompts.
Here, we provide the lists of prompts that were used for the experiments.

For singular form, below 15 templates were used:\newline
'A photo of a \{\}.',\newline
'A photo of a small \{\}.',\newline
'A photo of a medium \{\}.',\newline
'A photo of a large \{\}.',\newline
'This is a photo of a \{\}.',\newline
'This is a photo of a small \{\}.',\newline
'This is a photo of a medium \{\}.',\newline
'This is a photo of a large \{\}.',\newline
'A \{\} in the scene.',\newline
'A photo of a \{\} in the scene.',\newline
'There is a \{\} in the scene.',\newline
'There is the \{\} in the scene.',\newline
'This is a \{\} in the scene.',\newline
'This is the \{\} in the scene.',\newline
'This is one \{\} in the scene.',\newline

For plural form, below 11 templates were used:\newline
'A photo of a number of \{\}.'\newline
'A photo of a number of small \{\}.'\newline
'A photo of a number of medium \{\}.'\newline
'A photo of a number of large \{\}.'\newline
'There is a photo of a number of \{\}.'\newline
'There is a photo of a number of small \{\}.'\newline
'There is a photo of a number of medium \{\}.'\newline
'There is a photo of a number of large \{\}.'\newline
'A number of \{\} in the scene.'\newline
'A photo of a number of \{\} in the scene.'\newline
'There are a number of \{\} in the scene.'\newline

\section{Additional Qualitative Results}
\begin{figure*}[t]
    \begin{center}
        \includegraphics[width=\linewidth]{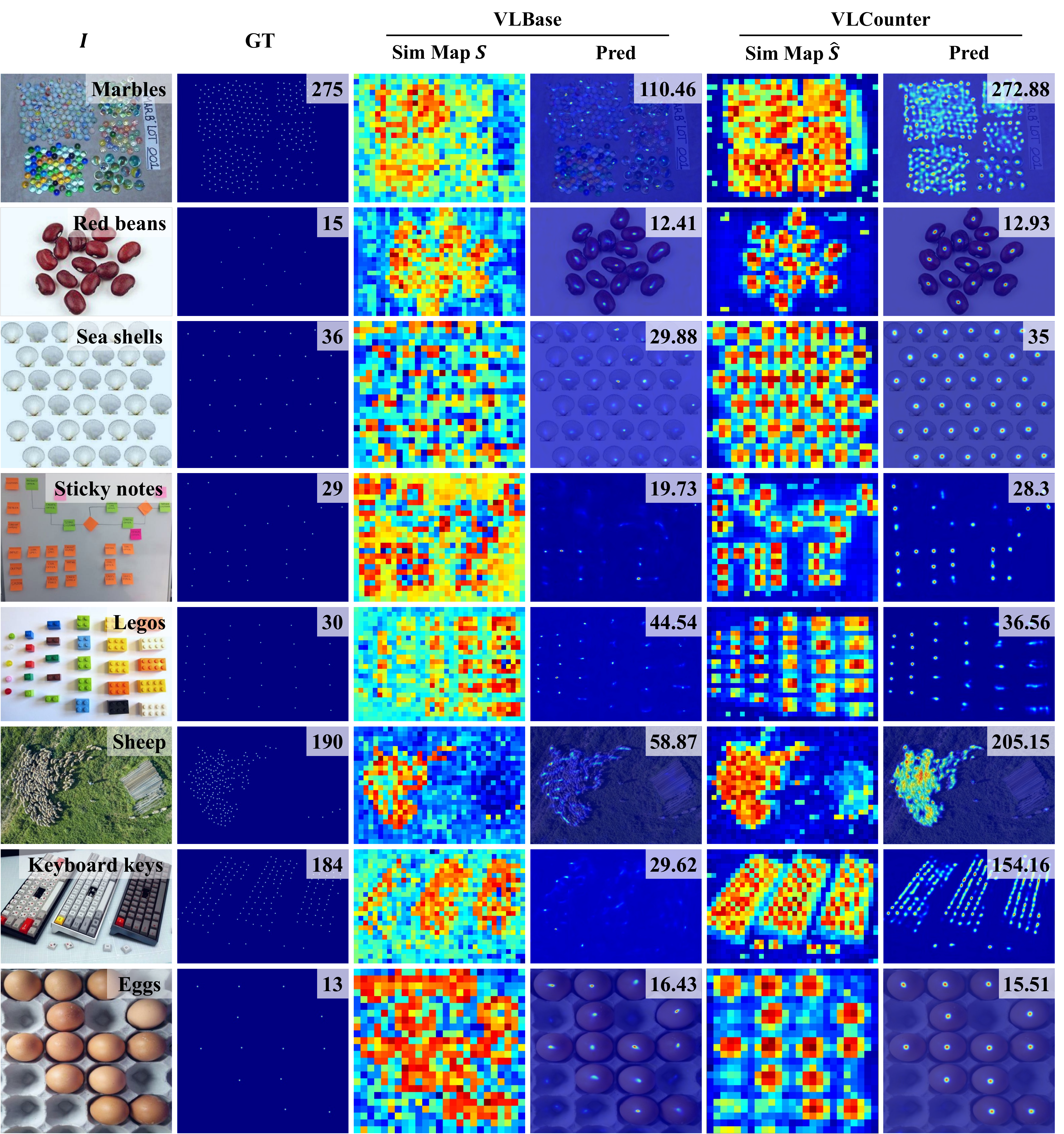}
    \end{center}
    \caption{
        Qualitative comparison of VLBase and VLCounter on the FSC-147.
        Class names and counting values are shown at the right top of the query image~($I$) and the predicted density map, respectively.
    }
    \label{fig:qualitative_results_suppl}
\end{figure*}

In addition to qualitative results in the manuscript, we provide more results in Fig.~\ref{fig:qualitative_results_suppl}, comparing the vision-language similarity map and the density map produced by our VLBase and VLCounter on the FSC147 dataset.
Note that we could not compare with the previous two-stage baselines since their implementations are not fully publicized.

% 정성 평가 결과 그림

\section{Comparison to Concurrent Work}
Recently, many efforts have been made to perform pixel-level dense prediction using CLIP.
While a concurrent work, CLIP-count~\cite{jiang2023clip}, requires additional parameters for visual-text interaction layers, we point out that our approach does not charge much memory cost since we leverage the semantic tokens within the image encoding process.
In Tab.~\ref{tab:clipcount}, we compare the number of learnable parameters and Multiply–ACcumulate~(MACs), revealing that our method shows an advantage in computational efficiency.
Moreover, while our performances seem to bring marginal benefits over CLIP-Count on the FSC-147 dataset~(Tab.~\ref{tab:clipcount_fsc}), we emphasize the large performance gaps in cross-domain scenarios in Tab.~\ref{tab:clipcount}~(+44.4\% and +5\% on CARPK and IOCfish5k datasets in MAE, respectively).

\begin{table}[h]
    \small
    \setlength{\extrarowheight}{2.3pt}
    \setlength{\tabcolsep}{1.5pt}
    \centering
    \begin{tabular*}{\linewidth}{l@{\extracolsep{\fill}}*{5}{c}}
    \hline
    \multicolumn{2}{c}{\multirow{2}{*}{Methods}} & \multicolumn{2}{c}{Val set} & \multicolumn{2}{c}{Test set} \\
    \cline{3-4}\cline{5-6}
     & & MAE & RMSE & MAE & RMSE \\
    \hline
    \multicolumn{2}{c}{CLIP-Count} & 18.76 & \textbf{61.18} & 17.78 & 106.62 \\
    \multicolumn{2}{c}{VLCounter~(Ours)} & \textbf{18.06} & 65.13 & \textbf{17.05} & \textbf{106.16} \\
    \hline
    % \hline
    \end{tabular*}
    \caption{Comparision with CLIP-Count on FSC147 dataset}
    \label{tab:clipcount_fsc}
\end{table}

\begin{table*}[t!]
    \small
    \setlength{\extrarowheight}{1.8pt}
    \centering
    % \begin{tabular}{ccccccc}
    \begin{tabular*}{\textwidth}{@{\extracolsep{\fill}}*{7}{c}}
    \hline
    Methods & Learnable Params~(M) & MACs~(G) & CARPK~(MAE) & CARPK~(RMSE) & IOCfish5k~(MAE) & IOCfish5k~(RMSE) \\
    \hline
    CLIP-Count & 16.36 & 123.06 & 11.70 & 13.94 & 82.1 & 155.2 \\
    \hline
    Ours & \textbf{1.44} & \textbf{34.36} & \textbf{6.46} & \textbf{8.68} & \textbf{78.0} & \textbf{154.9} \\
    \hline
    \end{tabular*}
    \caption{Comparision with CLIP-Count in the number of learnable parameters, MACs, and performance on diverse datasets}
    \label{tab:clipcount}
\end{table*}

\end{document}